\documentclass[conference,onecolumn]{IEEEtran}
\IEEEoverridecommandlockouts
\usepackage{siunitx}
\usepackage[acronym]{glossaries}
\usepackage{ifthen}
\usepackage{amsmath,amssymb,amsfonts}
\usepackage{algorithmic}
\usepackage{graphicx}
\usepackage{textcomp}
\usepackage{xcolor}
\usepackage{colortbl}
\usepackage{ragged2e}
\usepackage{tabularx}
\usepackage[bookmarks=false]{hyperref}
\usepackage{comment}
\usepackage{booktabs}
\usepackage{multirow}
\usepackage{dirtytalk}
\usepackage{tikz}
\usetikzlibrary{shapes.geometric, arrows, positioning}
\usepackage{circuitikz}
\usepackage{svg}
\usepackage{pgfplots}
\usepgfplotslibrary{fillbetween}
\pgfplotsset{width=10cm,compat=1.9}
\usepackage[most]{tcolorbox}
\usepackage[frozencache,cachedir=.]{minted}
\usepackage{enumitem}
\usepackage{caption}
\usepackage{listings}
\usepackage{wrapfig}
\usepackage{subcaption}
\usepackage{pifont}
\usepackage{soul}
\usepackage{footnote}
\usepackage{algorithm}
\usepackage{algorithmic}
\usepackage{pgfplotstable}

\definecolor{ForestGreen}{RGB}{34,150,34}

\newtcolorbox{bluebox}[2][]{
arc=4mm,
lower separated=false,
fonttitle=\bfseries,
colbacktitle=blue!40,
coltitle=black,
enhanced,
attach boxed title to top left={xshift=0.5cm,
        yshift=-2mm},
colframe=gray,
colback=white,
title=#2,#1}

\newtcolorbox{greenbox}[2][]{
arc=4mm,
lower separated=false,
fonttitle=\bfseries,
colbacktitle=teal!50,
coltitle=black,
enhanced,
attach boxed title to top left={xshift=0.5cm,
        yshift=-2mm},
colframe=gray,
colback=white,
title=#2,#1}

\makeatletter
\newcommand{\linebreakand}{%
\end{@IEEEauthorhalign}
\hfill\mbox{}\par
\mbox{}\hfill\begin{@IEEEauthorhalign}
}
\makeatother

\newboolean{blindreview}
\setboolean{blindreview}{false}
\DeclareRobustCommand{\IEEEauthorrefmark}[1]{\smash{\textsuperscript{\footnotesize #1}}}

\newacronym{SoC}{SoC}{System-on-Chip}
\newacronym{CWEs}{CWEs}{Common Weakness Enumerations}
\newacronym{CWE}{CWE}{Common Weakness Enumeration}
\newacronym{CVEs}{CVEs}{Common Vulnerability Enumerations}
\newacronym{CVE}{CVE}{Common Vulnerability Enumeration}
\newacronym{AI}{AI}{Artificial Intelligence}
\newacronym{ML}{ML}{Machine Learning}
\newacronym{LLMs}{LLMs}{Large Language Models}
\newacronym{LLM}{LLM}{Large Language Model}
\newacronym{ASIC}{ASIC}{Application Specific Integrated Circuit}
\newacronym{DUV}{DUV}{Design Under Verification}
\newacronym{CEX}{CEX}{Counter Example}
\newacronym{CEXs}{CEXs}{Counter Examples}
\newacronym{DOS}{DOS}{Denial of Service}
\newacronym{FSM}{FSM}{Finite State Machine}
\newacronym{FSMs}{FSMs}{Finite State Machines}
\newacronym{SEU}{SEU}{Single Event Upset}
\newacronym{SEUs}{SEUs}{Single Event Upsets}
\newacronym{RTL}{RTL}{Register Transfer Level}
\newacronym{IP}{IP}{Intellectual Property}
\newacronym{IPs}{IPs}{Intellectual Properties}
\newacronym{NLP}{NLP}{Natural Language Processing}
\newacronym{HDL}{HDL}{Hardware Description Language}
\newacronym{GPT}{GPT}{Generative Pre-trained Transformer}
\newacronym{PPA}{PPA}{Power, Performance and Area}
\newacronym{SVA}{SVA}{SystemVerilog Assertion}
\newacronym{SVAs}{SVAs}{SystemVerilog Assertions}
\newacronym{FV}{FV}{Formal Verification}
\newacronym{CSV}{CSV}{Comma-Separated Values}
\newacronym{RQ}{RQ}{Research Question}
\newacronym{RQs}{RQs}{Research Questions}
\newacronym{GenAI}{GenAI}{Generative AI}
\newacronym{UVM}{UVM}{Universal Verification Methodology}
\newacronym{ADHD}{ADHD}{Attention Deficit Hyperactivity Disorder}
\newacronym{FIFO}{FIFO}{First-In First-Out}
\newacronym{CoT}{CoT}{Chain-of-Thought}
\newacronym{MMLU}{MMLU}{Massive Multitask Language Understanding}
\newacronym{vPlan}{vPlan}{Verification Plan}
\newacronym{KPI}{KPI}{Key Performance Indicator}
\newacronym{KPIs}{KPIs}{Key Performance Indicators}
\newacronym{AGI}{AGI}{Artificial General Intelligence}
\newacronym{ALU}{ALU}{Arithmetic Logic Unit}
\newacronym{HIL}{HIL}{Human-in-the-Loop}

\usepackage[style=ieee, backend=biber, natbib=true, maxbibnames=1, minbibnames=1]{biblatex}
\begin{document}

\lstset{
    language=Verilog,           
    basicstyle=\footnotesize,   
    numbers=left,               
    frame=lines,                
    captionpos=b,               
    breaklines=true,            
    tabsize=2,                  
    xleftmargin=2.1em,
    framexleftmargin=1.7em,
    commentstyle=\color{ForestGreen},
    keywordstyle=\color{blue},
    stringstyle=\color{red},
}

\lstdefinelanguage{Verilog}{morekeywords={accept_on,alias,always,always_comb,always_ff,always_latch,and,assert,assign,assume,automatic,before,begin,bind,bins,binsof,bit,break,buf,bufif0,bufif1,byte,case,casex,casez,cell,chandle,checker,class,clocking,cmos,config,const,constraint,context,continue,cover,covergroup,coverpoint,cross,deassign,default,defparam,design,disable,dist,do,edge,else,end,endcase,endchecker,endclass,endclocking,endconfig,endfunction,endgenerate,endgroup,endinterface,endmodule,endpackage,endprimitive,endprogram,endproperty,endspecify,endsequence,endtable,endtask,enum,event,eventually,expect,export,extends,extern,final,first_match,for,force,foreach,forever,fork,forkjoin,function,generate,genvar,global,highz0,highz1,if,iff,ifnone,ignore_bins,illegal_bins,implements,implies,import,incdir,include,initial,inout,input,inside,instance,int,integer,interconnect,interface,intersect,join,join_any,join_none,large,let,liblist,library,local,localparam,logic,longint,macromodule,matches,medium,modport,module,nand,negedge,nettype,new,nexttime,nmos,nor,noshowcancelled,not,notif0,notif1,null,or,output,package,packed,parameter,pmos,posedge,primitive,priority,program,property,protected,pull0,pull1,pulldown,pullup,pulsestyle_ondetect,pulsestyle_onevent,pure,rand,randc,randcase,randsequence,rcmos,real,realtime,ref,reg,reject_on,release,repeat,restrict,return,rnmos,rpmos,rtran,rtranif0,rtranif1,s_always,s_eventually,s_nexttime,s_until,s_until_with,scalared,sequence,shortint,shortreal,showcancelled,signed,small,soft,solve,specify,specparam,static,string,strong,strong0,strong1,struct,super,supply0,supply1,sync_accept_on,sync_reject_on,table,tagged,task,this,throughout,time,timeprecision,timeunit,tran,tranif0,tranif1,tri,tri0,tri1,triand,trior,trireg,type,typedef,union,unique,unique0,unsigned,until,until_with,untyped,use,uwire,var,vectored,virtual,void,wait,wait_order,wand,weak,weak0,weak1,while,wildcard,wire,with,within,wor,xnor,xor,`uvm_create, `uvm_rand_send_with},morecomment=[l]{//}}

\title{Saarthi: The First AI Formal Verification Engineer\\

}

\ifthenelse{\boolean{blindreview}}{}{
	\author{\IEEEauthorblockN{
	      Aman Kumar\IEEEauthorrefmark{1},
            Deepak Narayan Gadde\IEEEauthorrefmark{2},
			Keerthan Kopparam Radhakrishna\IEEEauthorrefmark{2},
			Djones Lettnin\IEEEauthorrefmark{3}}
		\IEEEauthorblockA{
			\IEEEauthorrefmark{1}Infineon Technologies Semiconductor India Private Limited, India \\
            \IEEEauthorrefmark{2}Infineon Technologies Dresden GmbH \& Co. KG, Germany \\
            \IEEEauthorrefmark{3}Infineon Technologies AG, Germany}
	}
}

\maketitle

\begin{abstract}
\textbf{Recently, Devin has made a significant buzz in the \acrfull{AI} community as the world’s first fully autonomous AI software engineer, capable of independently developing software code \cite{devin} \cite{opendevin}. Devin uses the concept of agentic workflow in \acrfull{GenAI}, which empowers AI agents to engage in a more dynamic, iterative, and self-reflective process. In this paper, we present a similar fully autonomous AI formal verification engineer, Saarthi\footnote{Saarthi is a Sanskrit word that means someone who guides and leads you to your destination.}, capable of verifying a given \acrshort{RTL} design end-to-end using an agentic workflow. With Saarthi, verification engineers can focus on more complex problems, and verification teams can strive for more ambitious goals. The domain-agnostic implementation of Saarthi makes it scalable for use across various domains such as \acrshort{RTL} design, \acrshort{UVM}-based verification, and others.}
\end{abstract}


\section{Introduction}

Hardware design verification, especially formal verification, entails a methodical and disciplined approach to the planning, development, execution, and sign-off of functionally correct hardware designs. Formal verification uses mathematical methods to prove the correctness of hardware designs against their specifications, ensuring that all possible states and inputs are considered, which complements traditional simulation-based verification techniques that might only cover a subset of possible scenarios due to practical constraints \cite{fvbook}. The formal verification process encompasses several key roles, such as organizational coordination, task allocation, code development, property proving, analyzing \acrfull{CEXs}, debugging, coverage closure, and documentation preparation. These roles are crucial for managing the complexity and ensuring the thoroughness of the verification process. For instance, analyzing counterexamples involves identifying specific scenarios where the design might fail to meet its specifications, which is critical for debugging and refining the design. This highly intricate activity demands meticulous attention to detail, given its long development cycles and the critical nature of ensuring hardware functionality and reliability \cite{VerStudy}.

The field of \acrfull{NLP} has undergone a significant transformation with the advent of \acrfull{LLMs} \cite{deepseek}. These powerful models, often referred to as \acrshort{GenAI}, have revolutionized how machines understand and generate human language, enabling unprecedented advancements in a wide array of applications \cite{attention}. Through extensive training on large datasets using the \say{next word prediction} approach, LLMs have demonstrated remarkable capabilities in various downstream tasks such as context-sensitive question answering, machine translation, and code generation \cite{chatdev}. Interestingly, the primary components of formal verification -- specifically code (assertions as properties) and specification documents -- can be considered as forms of \say{language} or sequences of characters \cite{llm_lang}. Various surveys \cite{fv_survey} have discussed techniques for improving conventional formal verification; however, we aim to enhance it further using \acrshort{AI}. This paper introduces an end-to-end formal verification process driven by \acrshort{LLMs}. This process encompasses design specification analysis, code development, verification, and document generation, aiming to establish a unified, efficient, and cost-effective paradigm for hardware design verification. By leveraging the advanced capabilities of \acrshort{LLMs}, it is possible to streamline the formal verification process, enhancing both accuracy and productivity.

Like every other semiconductor company, we wanted to investigate the possibilities of using \acrshort{GenAI} for dedicated use cases. However, there are known challenges that prevent precise use case definitions \cite{erik}. \acrshort{GenAI} operates as a stochastic process, meaning it generates non-deterministic output with each regeneration. This is in contrast to the requirements of hardware design verification, which demands precise, deterministic answers, particularly in formal verification where engineers need to make clear pass/fail decisions based on exact criteria. Due to this fundamental mismatch, the non-deterministic output of \acrshort{LLMs} is not always suitable for hardware design verification. Additionally, \acrshort{LLMs} can suffer from artificial \acrfull{ADHD}, characterized by a tendency to lose focus on the task at hand, and hallucination, where the model generates incorrect or nonsensical information confidently \cite{reformai}. These issues are very prominent and can lead \acrfull{GPT} users to get stuck in iterative loops, repeatedly seeking accurate results without success. Given these challenges, the current state of \acrshort{GenAI} may not be well-suited for applications that require the high precision and determinism essential in hardware design verification.

To overcome the aforementioned challenges and use \acrshort{GenAI} for problem-solving, we introduce a fully autonomous AI formal verification engineer, Saarthi, capable of verifying a given \acrshort{RTL} design end-to-end using an agentic workflow. Saarthi stands for \underline{S}cal\underline{A}ble \underline{ART}ificial and \underline{H}uman \underline{I}ntelligence that uses agentic \acrshort{AI} based reasoning patterns and human intelligence to produce sensible results. Saarthi can run formal verification on several complex \acrshort{IPs} with coverage closure and find bugs similar to a human verification engineer. Our contributions to this work are summarized below:

\begin{itemize}
    \item We propose Saarthi, an agentic \acrshort{AI}-based formal verification engineer. By providing the design specification, Saarthi sequentially handles verification planning, generating \acrfull{SVAs}, proving the properties, analyzing \acrshort{CEXs}, and analyzing the formal coverage for sign-off.
    \item To address the issue of \acrshort{ADHD} and code hallucination, we use the approach of agentic (few-shot) workflow as opposed to the non-agentic (zero-shot) workflow.
    \item To further alleviate potential challenges related to \acrshort{LLMs} getting stuck in iterative loops, we use the concept of \say{\acrfull{HIL}} \acrshort{AI} to ensure uninterrupted end-to-end formal verification.
\end{itemize}

Section \ref{bgnd} summarises the related work and introduces agentic workflow design patterns. Section \ref{setup} discusses the details about the framework of \acrshort{AI} agents and how they interact with each other to perform formal verification. Section \ref{eval} presents our results from evaluating formal verification of different \acrshort{RTL} designs with varied complexity. Section \ref{conc} concludes the paper with an outlook on possible future research opportunities.

\section{Background} \label{bgnd}

Recent work in \cite{reformai} found that \acrshort{LLMs} tend to generate incorrect designs and are vulnerable to security flaws as the authors observed around \SI{60}{\percent} failure rate for the generated \acrshort{RTL} designs. Our findings indicated that the expectations of authors from \acrshort{AI} were too high in terms of achieving first-time-right designs. Additionally, their approach of benchmarking the LLM-generated designs may have been overly pessimistic, possibly not accounting for the iterative improvement potential of these models. Moving forward, we want to focus on using \acrshort{GenAI} as a problem-solving tool and to use the existing capabilities of \acrshort{LLMs} to generate better results. There are two types of \acrshort{AI}-based workflows:

\begin{itemize}
    \item Non-agentic workflow (zero-shot)
    \item Agentic workflow (few-shot)
\end{itemize}

\subsection{Non-Agentic Workflow}

\begin{wrapfigure}{r}{0.25\textwidth}
  \centering
    \includegraphics[width=0.5\linewidth]{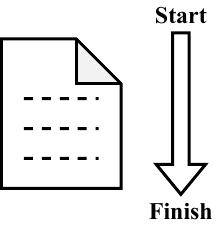}
  \caption{Non-agentic workflow}
  \label{zero_shot}
\end{wrapfigure}
The first productive uses of \acrshort{LLMs} involved non-agentic workflows, where we type a prompt and the model generates an answer in one go. This is akin to asking a person to write an essay on a topic and saying \say{please sit down to the keyboard and type the essay from start to finish without ever using backspace}. Despite how hard this is, \acrshort{LLMs} do it remarkably well; however, the quality of the generated content is often relatively lower due to the lack of iterative refinement. This approach is termed a zero-shot or non-agentic workflow. In a zero-shot scenario, the model attempts to generate a response without any prior specific examples or iterations tailored to the task at hand.

In the ReFormAI dataset paper \cite{reformai}, the authors used a similar non-agentic workflow and benchmarked the \acrshort{LLM} generated \acrshort{RTL} codes that resulted in a relatively higher failure rate. The results suggested that the failure rate was significant due to the one-pass, zero-shot nature of the generation process. The results would likely have been better if a feedback loop had been added to the generation part. A feedback loop would allow for iterative refinement, where the model could receive feedback on its initial outputs and make adjustments to improve accuracy and quality. This approach would enable the \acrshort{LLM} to correct errors, incorporate additional context, and ultimately produce higher-quality \acrshort{RTL} designs.

\subsection{Agentic Workflow}

\begin{wrapfigure}{r}{0.25\textwidth}
  \centering
    \includegraphics[width=0.7\linewidth]{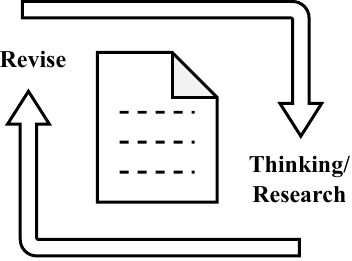}
  \caption{Agentic workflow}
  \label{few_shot}
\end{wrapfigure}
In contrast to the zero-shot workflow, the agentic or few-shot workflow uses iterative loops and feedback to produce better results. This approach is very similar to how humans think and approach a given task. For the task of writing an essay, a human would typically start by outlining the essay on topic X, conducting web research, preparing a first draft, considering what parts of the essay need revision, revising the draft, and finally producing the final version. Similarly, if \acrshort{LLMs} employ this iterative approach to address a prompt, they deliver remarkably better results. In a few-shot workflow, the model is initially provided with a few examples to guide its responses. As it generates outputs, it receives feedback, which it uses to refine and improve its responses iteratively. This process allows for error correction and the incorporation of additional context, leading to higher quality and more accurate results compared to the zero-shot approach.

Based on open-source benchmarks, researchers found that using OpenAI models \cite{gpt4o} such as \acrshort{GPT}-3.5 with zero-shot prompting, the LLM yields \SI{48}{\percent} correct results. With \acrshort{GPT}-4, this accuracy improves to \SI{67}{\percent}. However, when using an agentic workflow and wrapping it around \acrshort{GPT}-3.5, it outperformed \acrshort{GPT}-4, demonstrating the significant impact of iterative feedback and refinement \cite{youtube}. Certainly, an agentic workflow wrapped around \acrshort{GPT}-4 produced even better results, further enhancing accuracy and performance. We also tried a similar approach to formally verify a synchronous \acrshort{FIFO} and achieved a \SI{100}{\percent} pass rate using the few-shot approach within minutes. Table \ref{fifo_few_shot} summarizes the result.

\begin{table}[ht]
\renewcommand\arraystretch{1.2}
\centering
\caption{Zero-shot and few-shot prompting for formal verification of a synchronous \acrshort{FIFO} design}
\begin{tabular}[t]{ccccc}
\toprule
\textbf{Workflow} & \textbf{Proved Assertions} & \textbf{\acrshort{CEX}} & \textbf{Unreachable Covers} & \textbf{Covered Covers}\\
\toprule
Zero-shot & \SI{42.85}{\percent} & \SI{57.15}{\percent} & \SI{12.5}{\percent} & \SI{87.5}{\percent}\\
Few-shot & \SI{100}{\percent} & \SI{0}{\percent} & \SI{0}{\percent} & \SI{100}{\percent}\\
\bottomrule
\end{tabular}
\label{fifo_few_shot}
\end{table}%

Researchers have recently put a lot of effort into defining agentic reasoning design patterns. The most significant ones that facilitate agentic AI-based workflows are:

\begin{itemize}
    \item Reflection \cite{self_refine} \cite{reflexion}
    \item Tool use \cite{llm_api} \cite{mmreact}
    \item Planning \cite{chain_of_thought} \cite{hugging_gpt}
    \item Multi-agent collaboration \cite{chatdev} \cite{autogen}
\end{itemize}

\subsection{Reflection}

Reflection uses the concept of a coder agent and a critic agent. For any given task, there would be a coder agent that generates code—in our case, \acrfull{SVA} -- and a critic agent that critically analyzes and reviews the output of the coder agent, providing feedback. This feedback loop is iterative, allowing the coder agent to refine its code based on the critic agent's insights, leading to progressively better results.

\begin{figure}[h!]
\centering
  \includegraphics [width=0.70\textwidth] {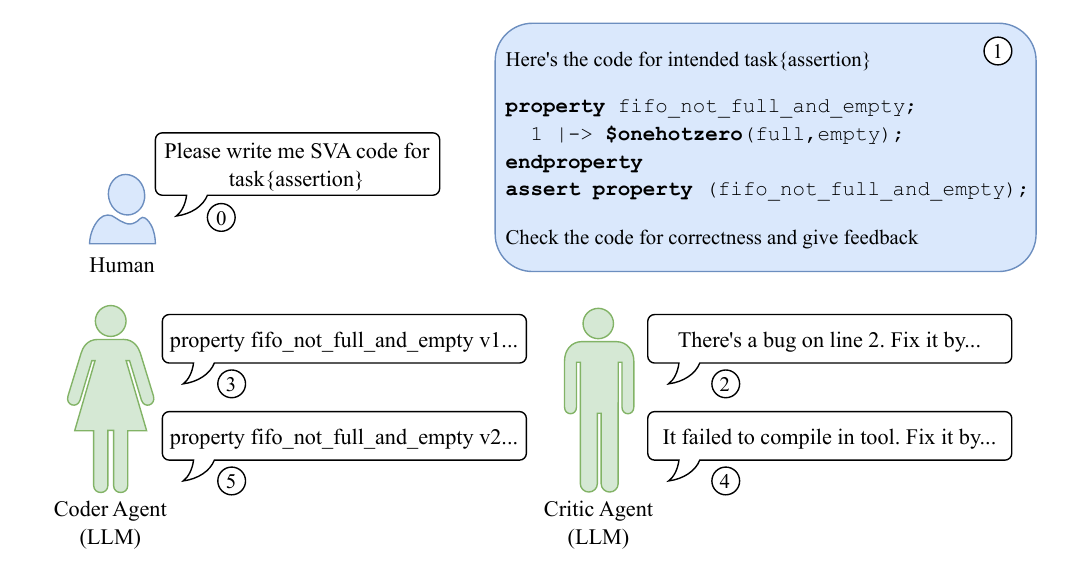}
\caption{Coder and critic \acrshort{AI} agents for self-reflection (adapted from \cite{youtube})}
\label{reflection}
\end{figure}

Fig.~\ref{reflection} shows a case where a human asked the coder \acrshort{AI} agent to write \acrshort{SVA} code for a given specification. Once the \acrshort{SVA} is generated, the critic agent analyzes the code and provides feedback, identifying a bug in line X. The coder agent then uses this feedback to fix the code and generate version 1 (v1) of the \acrshort{SVA}. Next, the critic agent attempts to compile the generated code using a formal verification tool and reports a compilation issue, including the error message from the tool, to the coder agent. The coder agent analyzes this feedback and produces a corrected \acrshort{SVA} as version 2 (v2). Using this iterative approach, the human is able to obtain a significantly better \acrshort{SVA} by leveraging the capabilities of existing LLMs. The human's role includes initiating the process, reviewing the iterations, and making use of the final, refined \acrshort{SVA} code.

\acrshort{LLM} agents are increasingly being used to interact with external environments as goal-driven agents. However, these language agents face difficulties in rapidly and effectively learning through trial-and-error, since conventional reinforcement learning techniques necessitate a large number of training samples and expensive model fine-tuning. The authors in \cite{reflexion} propose a novel framework, Reflexion, that uses verbal reinforcement to help agents learn from previous failures. Creating valuable reflective feedback is difficult because it involves accurately identifying where the model went wrong (known as the credit assignment problem \cite{reinforcement_learning}) and being able to produce a summary that offers actionable recommendations for improvement. The authors also propose several mitigation techniques such as simple binary environment feedback, pre-defined heuristics for common failure cases, and self-evaluation such as binary classification using \acrshort{LLMs} (decision-making) or self-written unit tests (programming).

\begin{figure}[h!]
    \centering
    \begin{minipage}{.3\textwidth}
        \centering
        \includegraphics[width=0.9\textwidth]{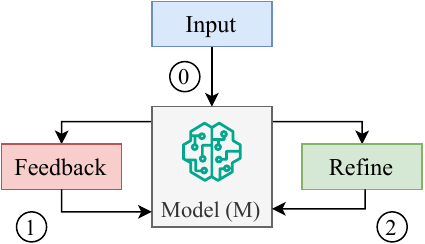}
        \caption{Iterative feedback with self-refinement \cite{self_refine}}
        \label{self_refine}
    \end{minipage}%
    \hfill
    \begin{minipage}{0.6\textwidth}
        \centering
        \includegraphics[width=0.95\textwidth]{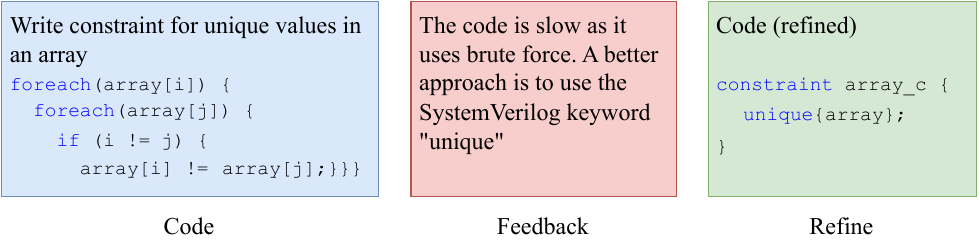}
        \caption{Example of self-refinement: an initial output is generated by the base \acrshort{LLM} and then passed back to the same \acrshort{LLM} to receive feedback and sent to the same \acrshort{LLM} to refine the output.}
        \label{self_refine_example}
    \end{minipage}
\end{figure}

The authors in \cite{self_refine} introduce iterative self-refinement, a fundamental characteristic of human problem-solving that involves creating an initial draft and subsequently refining it based on self-provided feedback. A similar approach can be applied to \acrshort{LLM} agents as shown in Fig.~\ref{self_refine}. Given an input (0), self-refinement starts by generating an output and passing it back to the same model \textit{M} to get feedback (1). Feedback is passed back to \textit{M}, which refines the previously generated output (2). Steps (1) and (2) iterate until a stopping condition is met. An example of such a self-refinement is highlighted in Fig.~\ref{self_refine_example}. Using self-feedback, the \acrshort{LLM} agent could modify the previously generated code for a constraint in SystemVerilog to generate unique values in a random array to an optimized version.

\subsection{Tool Use}

Tool use is another agentic reasoning design pattern that leverages \acrshort{AI}-based tools to generate meaningful results. Web search tools such as Copilot, which assists with coding by providing intelligent code suggestions, and code execution tools such as \cite{gorilla} and \cite{mmreact}, which help automate and streamline the coding process, significantly increase productivity gains. These tools enable users to accomplish tasks more efficiently by harnessing the power of \acrshort{AI} to provide relevant information, automate repetitive processes, and enhance overall workflow. Some examples of areas of application for tool use are mentioned in Table \ref{tool_use}.

\begin{table}[ht]
\renewcommand\arraystretch{1.2}
\centering
\caption{Tool use in agentic workflow \cite{youtube}}
\begin{tabular}{llll}
\toprule
\textbf{Analysis}                & \textbf{Research}      & \textbf{Productivity}  & \textbf{Images}                           \\
\toprule
Code Execution          & Search engine & Email         & Imager generation (e.g., Dall-E) \\
Wolfram Alpha           & Web browsing  & Calendar      & Image captioning                  \\
Bearly Code Interpreter & Wikipedia     & Cloud Storage & Object detection                \\
\bottomrule
\end{tabular}
\label{tool_use}
\end{table}

\begin{figure}[h!]
\centering
\begin{tcbraster}[raster columns=2,raster equal height]
\begin{bluebox}[]{\footnotesize{Input Prompt}}
\footnotesize{You are a helpful \acrshort{AI} assistant that can help analyze the \acrfull{CEXs} from formal verification runs. Please analyze and explain the failure in the given \acrshort{CEX}.

\begin{center}
\includegraphics [width=0.3\textwidth] {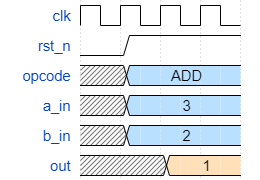}
\end{center}
}
\end{bluebox}
\begin{greenbox}[]{\footnotesize{Model Output}}
\footnotesize{As a helpful \acrshort{AI} assistant, I will first analyze the uploaded \acrshort{CEX} image...

\vspace{0.3cm}

After analyzing the \acrshort{CEX}, it appears that the opcode for \acrshort{ALU} is an addition operation and the inputs are 3 and 2. The addition of 3 and 2 should be 5 but the output is 1 which is wrong. The design is likely doing a subtraction instead of an addition.
}
\end{greenbox}
\end{tcbraster}
\caption{Using \acrshort{GenAI} tool to analyze a \acrshort{CEX}}
\label{cex_analysis_prompt}
\end{figure}

The authors in \cite{genai_sst} used \acrshort{GPT} tools to analyze inductive-step failure \acrshort{CEXs} in formal verification to generate helper assertions. A similar approach could be employed to analyze the \acrshort{CEXs} and request an explanation of the failure. Based on this analysis, the user can decide whether the \acrshort{RTL} design needs a fix or if the property should be adjusted in the case of a false-positive. An example of such a conversation between a human and the \acrshort{LLM} is highlighted in Fig.~\ref{cex_analysis_prompt}, demonstrating how the \acrshort{LLM} provides insights that aid the user in making these decisions.

\subsection{Planning}

\begin{figure}[h!]
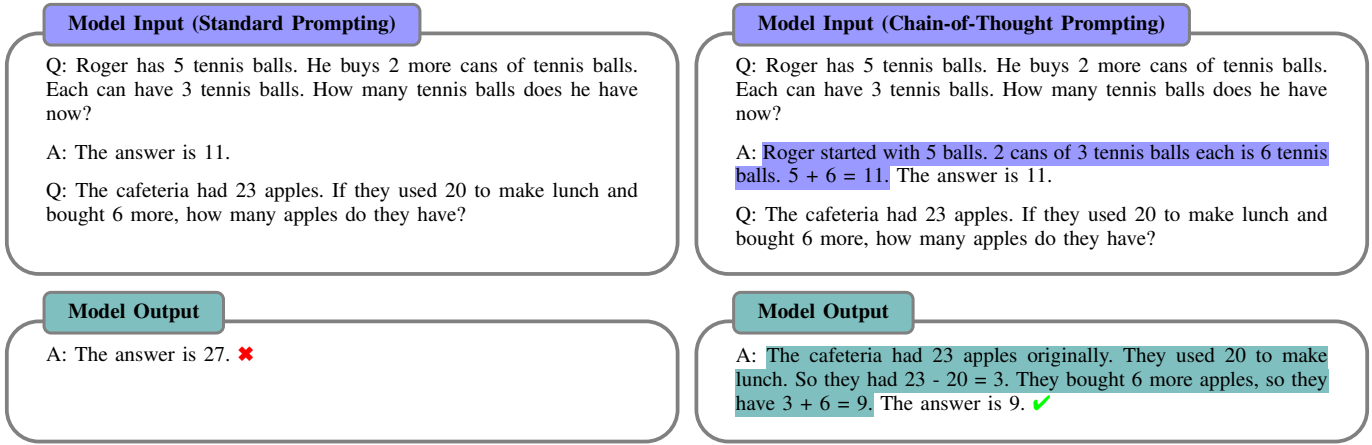

\centering
\begin{tcbraster}[raster columns=2,raster equal height]
\begin{bluebox}[]{\footnotesize{Model Input (Standard Prompting)}}
\footnotesize{Q: Roger has 5 tennis balls. He buys 2 more cans of tennis balls. Each can have 3 tennis balls. How many tennis balls does he have now?

\vspace{0.2cm}

A: The answer is 11.

\vspace{0.2cm}

Q: The cafeteria had 23 apples. If they used 20 to make lunch and bought 6 more, how many apples do they have?}
\end{bluebox}
\begin{bluebox}[]{\footnotesize{Model Input (Chain-of-Thought Prompting)}}
\footnotesize{Q: Roger has 5 tennis balls. He buys 2 more cans of tennis balls. Each can have 3 tennis balls. How many tennis balls does he have now?

\vspace{0.2cm}

A: \sethlcolor{blue!40}\hl{Roger started with 5 balls. 2 cans of 3 tennis balls each is 6 tennis balls. 5 + 6 = 11.} The answer is 11.

\vspace{0.2cm}

Q: The cafeteria had 23 apples. If they used 20 to make lunch and bought 6 more, how many apples do they have?}
\end{bluebox}
\end{tcbraster}

\begin{tcbraster}[raster columns=2,raster equal height]
\begin{greenbox}[]{\footnotesize{Model Output}}
\footnotesize{A: The answer is 27.} {\textcolor{red}{\ding{54}}}
\end{greenbox}
\begin{greenbox}[]{\footnotesize{Model Output}}
\footnotesize{A: \sethlcolor{teal!50}\hl{The cafeteria had 23 apples originally. They used 20 to make lunch. So they had 23 - 20 = 3. They bought 6 more apples, so they have 3 + 6 = 9.} The answer is 9.} {\textcolor{green}{\ding{52}}}
\end{greenbox}
\end{tcbraster}
\caption{Standard prompting vs. chain-of-thought prompting. Chain-of-thought reasoning processes are highlighted \cite{chain_of_thought}.}
\label{cot_prompting}
\end{figure}

Planning is one of the key workflows to generate quality output from \acrshort{LLMs} using agentic reasoning. Even though \acrshort{LLMs} are quite good at generating responses for a given prompt, they are often criticized for their lack of commonsense reasoning. This deficiency can impact the accuracy and reliability of the generated responses. Fig.~\ref{cot_prompting} illustrates this issue through a standard prompting example where the model's output is incorrect for a simple reasoning question. By incorporating planning into the workflow, \acrshort{LLMs} can be guided to consider broader context and logical steps, which helps mitigate the lack of commonsense reasoning and leads to higher quality outputs.

The authors in \cite{chain_of_thought} explore generating a \acrfull{CoT} -- a series of intermediate reasoning steps that enable \acrshort{LLMs} to tackle complex arithmetic, commonsense, and symbolic reasoning tasks. \acrshort{CoT}, in principle, allows models to decompose multi-step problems into intermediate steps, which means that additional computation can be allocated to problems that require more reasoning steps. For many reasoning tasks where standard prompting has a flat scaling curve, \acrshort{CoT} prompting leads to dramatically increasing scaling curves. An example of \acrshort{CoT} prompting is shown in Fig.~\ref{cot_prompting} that elicits reasoning in \acrshort{LLMs}.

Although \acrshort{CoT} emulates the thought process of human reasoners, this does not necessarily indicate that the neural network is actually \say{reasoning}. \acrshort{CoT} typically involves few-shot prompting, where the model is provided with a few examples to guide its responses. This approach can be expensive, especially when using paid \acrshort{LLMs}. In contrast, using zero-shot prompting with a more generalized prompt could be more cost-effective. Furthermore, there is no guarantee that \acrshort{CoT} will follow correct reasoning paths, which can lead to both correct and incorrect answers. The variability and uncertainty in the reasoning process mean that while \acrshort{CoT} can help generate more logically structured responses, it can also propagate errors if the initial reasoning path is flawed.

\subsection{Multi-Agent Collaboration}

\acrshort{AI} agents can collaborate to solve tasks given by a human. These agents can leverage several \acrshort{LLMs} to handle different responsibilities within a complex task. The authors in \cite{chatdev} introduce communicative agents for software development, which are designed to interact and share information to improve task-solving efficiency, and present an open-source alternative to Devin \cite{devin}. Research done in \cite{multiagent_debate} supports the notion that a multi-agent system performs better than a single agent when solving complex tasks. Table \ref{multiagent_debate_table} summarizes the results of a multi-agent debate for different tasks, such as the \acrfull{MMLU} benchmark and chess moves, demonstrating the improved performance of multi-agent systems in these diverse scenarios.

\begin{table}[ht]
\renewcommand\arraystretch{1.2}
\centering
\caption{Multi-agent debate \cite{multiagent_debate}}
\begin{tabular}[t]{ccc}
\toprule
\textbf{Task} & \textbf{Single Agent} & \textbf{Multi-Agent}\\
\toprule
Biographies & \SI{66.0}{\percent} & \SI{73.8}{\percent}\\
\acrshort{MMLU} & \SI{63.9}{\percent} & \SI{71.1}{\percent}\\
Chess move & \SI{29.3}{\percent} & \SI{45.2}{\percent}\\
\bottomrule
\end{tabular}
\label{multiagent_debate_table}
\end{table}%

A classic example of multi-agent collaboration is depicted in Fig. ~\ref{reflection}, where the coder and critic agents work together to solve a given task that includes reasoning and feedback. In this scenario, the coder agent generates the code, while the critic agent reviews the output and provides feedback to improve accuracy. To compensate for problems such as code hallucination -- where an AI generates plausible but incorrect code -- it is usually better to divide a complex task into simpler tasks and have one agent solve each of them \cite{more_agents}. This approach not only reduces the risk of hallucination but also mitigates the risk of an agent getting stuck in an iterative loop while attempting to solve a complex task. By breaking down the task, each agent can focus on a specific aspect, leading to more efficient and accurate problem-solving.

\begin{figure}[h!]
\centering
  \includegraphics [width=0.50\textwidth] {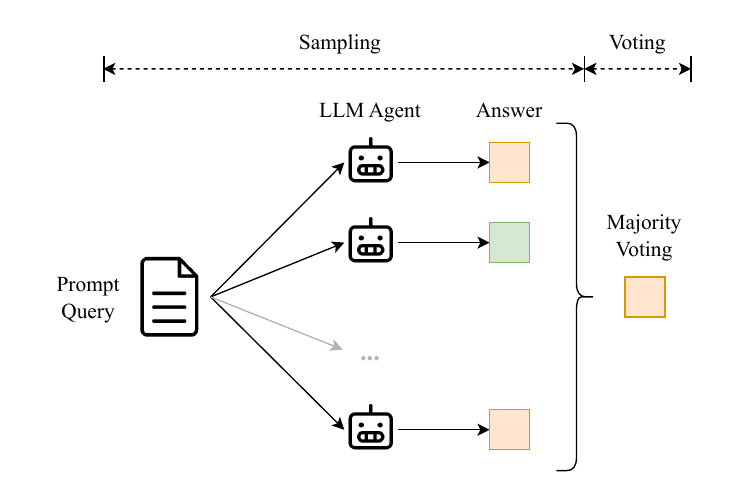}
\caption{Sampling-and-voting method \cite{more_agents}}
\label{sample_and_vote}
\end{figure}

The authors in \cite{more_agents} suggest the so-called \say{sampling-and-voting} method to improve results from multiple \acrshort{LLM} agents. This approach involves generating results from multiple \acrshort{LLM} agents for the same prompt (sampling) and then voting on the majority result to obtain the best possible outcome. This method leverages the diversity of responses to enhance accuracy and reliability. The performance of \acrshort{LLMs} scales with the number of agents instantiated, meaning that as more agents are used, the overall accuracy and quality of the results improve. This scaling effect is highlighted in Fig.~\ref{sample_and_vote}, which demonstrates how increasing the number of agents leads to better performance metrics such as accuracy, consistency, and robustness of the outputs.

\section{Saarthi: Agentic \acrshort{AI}-Based Formal Verification} \label{setup}

\begin{figure}[h!]
\centering
  \includegraphics [width=0.75\textwidth] {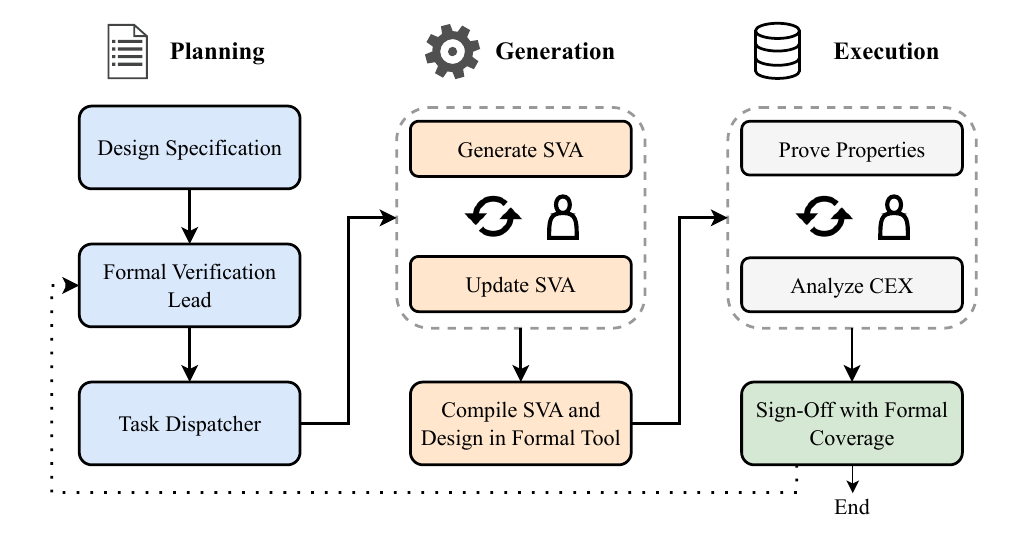}
\caption{Saarthi: Agentic \acrshort{AI} based formal verification using multi-agent collaboration}
\label{formal_flow}
\end{figure}

To realize our contributions and conduct our experiments, we prepared a flow as illustrated in Fig.~\ref{formal_flow} where \acrshort{AI} agents are in the driver's seat as soon as a task is given to solve. Saarthi is designed to facilitate formal verification through a sophisticated agentic \acrshort{AI}-based approach that leverages multi-agent collaboration. Saarthi integrates several design patterns, including agentic reasoning and techniques to mitigate issues such as attention deficits, hallucinations, and repetitive looping. It is built using three open-source frameworks -- CrewAI, AutoGen, and LangGraph -- enabling the user to select any of them for formal verification. The architecture implements a configurable agent orchestration system that can be adapted to varying verification requirements while maintaining consistency and reliability in the verification process.

\subsection{Flow Architecture}

\begin{figure}[h!]
\centering
  \includegraphics [width=0.89\textwidth] {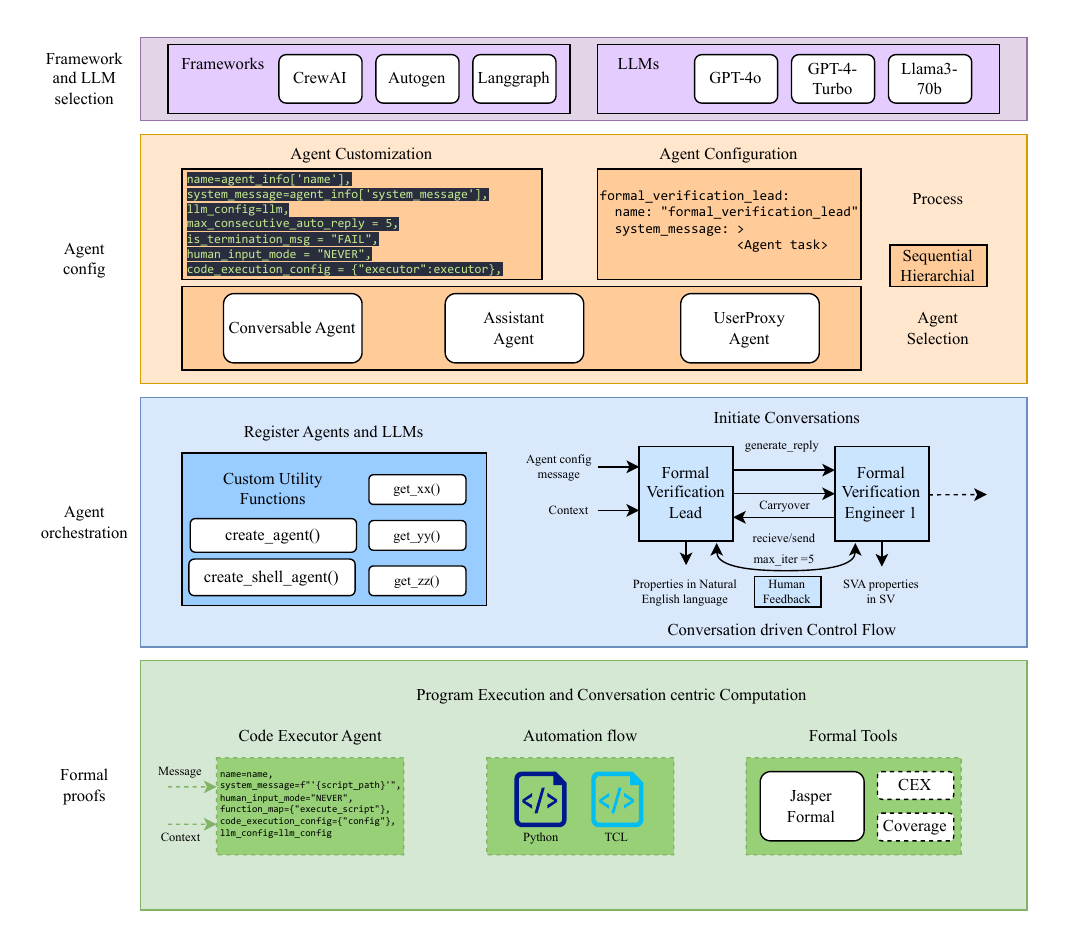}
\caption{Example usage of Saarthi for formal verification using multi-agent conversation}
\label{saaarthi_example}
\end{figure}

\begin{wrapfigure}{r}{0.28\textwidth}
  \centering
    \includegraphics[width=\linewidth]{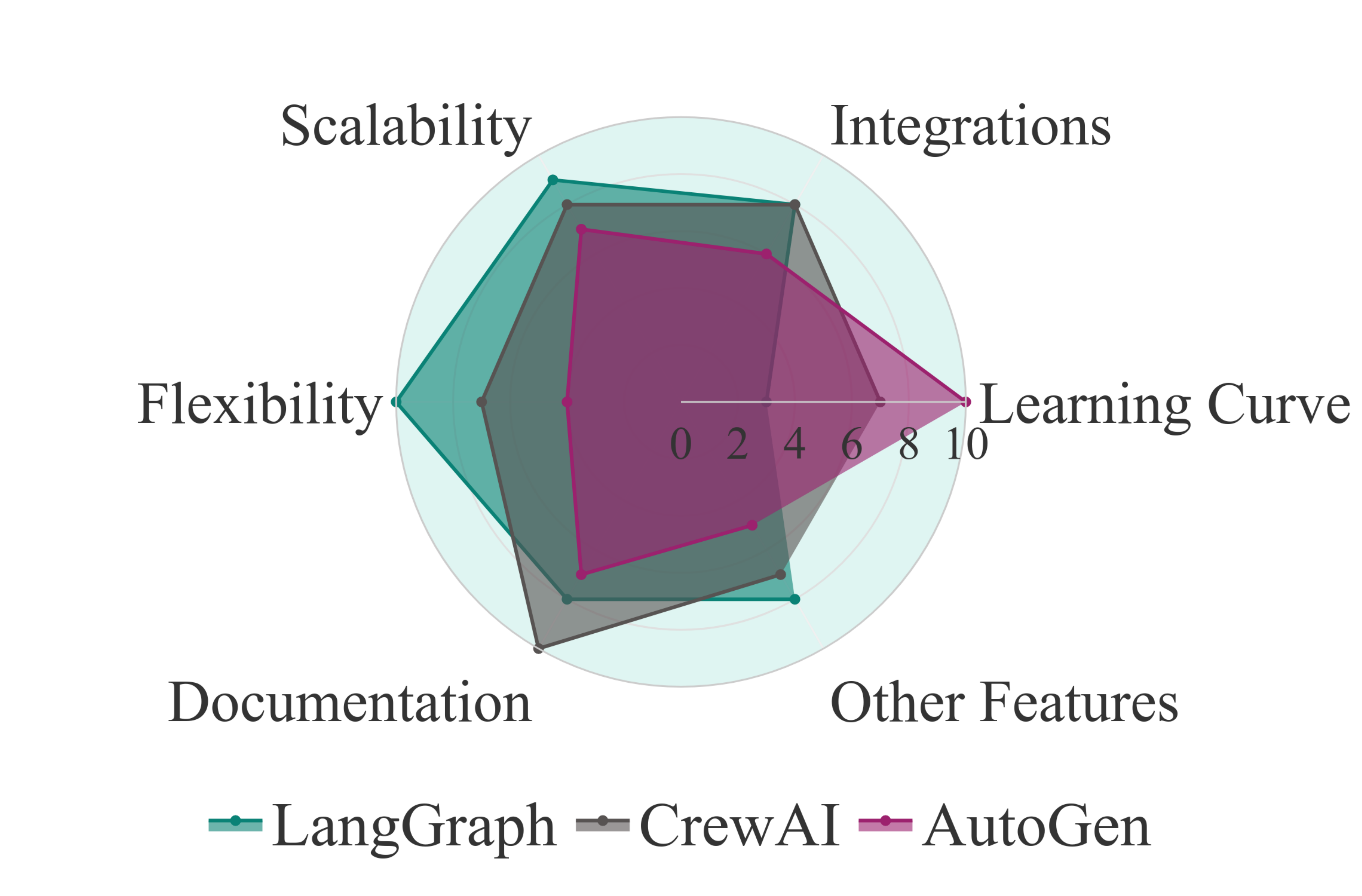}
  \caption{Comparison of different frameworks}
  \label{framework_comparison}
\end{wrapfigure}
The system's entry point is the main function, which implements an argument parsing system supporting multiple configuration domains, including framework selection mechanisms and \acrshort{LLM} model specifications. The framework selection encompasses implementations such as CrewAI, AutoGen, and LangGraph, while the \acrshort{LLM} configuration supports models including GPT-4o, GPT-4-Turbo, and Llama3-70B \cite{llama}. This modular architecture enables seamless switching between different implementation frameworks while maintaining consistent verification workflows. This main function also processes input files, which consist of a specification file for the input design. Each framework has an agent configuration file that defines the agents, including their roles and descriptions. The collaborative formal verification process is realized through a structured crew of agents, each specializing in a distinct aspect of verification. Fig.~\ref{framework_comparison} gives a comparative overview of the capabilities of different frameworks.

The AutoGen implementation features a sequential processing pipeline that manages inter-agent communication through structured message-passing protocols. The system implements comprehensive logging mechanisms that capture detailed interactions and logs. The CrewAI implementation utilizes decorator patterns for task definition, implementing a structured approach to workflow management. The system incorporates file-based tool integration mechanisms and implements a comprehensive logging system that captures execution details at multiple granularity levels.

The framework also implements an error management system incorporating multiple feedback loops for continuous improvement. The system utilizes critic agents that perform property evaluations, providing detailed feedback for assertion improvement. The iteration control system uses threshold-based monitoring to prevent infinite loops, automatically triggering human intervention when resolution cannot be achieved autonomously. The error-handling system implements structured exception management through try-catch hierarchies. Saarthi also generates verification artifacts through template engines and formatting systems. The logging system implements timestamp-based organization and multi-level capture, ensuring complete traceability of the verification process. Fig.~\ref{saaarthi_example} shows an example usage of Saarthi to accomplish a task using multi-agent collaboration.

Fig.~\ref{formal_flow} is the high-level overview of the flow we have defined for formal verification where \acrshort{AI} agents are in the driver's seat as soon as a task is given to solve. Every block in the flow chart is executed by an \acrshort{LLM}, except the first, the design specification that a human provides. The first \acrshort{LLM} agent is the so-called \say{formal verification lead} who is responsible for generating a \acrfull{vPlan} (i.e., a list of the properties necessary to verify the given design written in the natural English language) based on the given specification. The verification lead divides the follow-up tasks to other \acrshort{LLM} agents. An example of such an agent in the AutoGen framework with its role, goal and backstory is highlighted in Listing 1. The subsequent steps in the flow involve several \acrshort{LLMs} acting as formal verification engineers to analyze the \acrshort{vPlan} and generate \acrshort{SVA} for each corresponding element. An example of the tasks defined in the AutoGen framework for \acrshort{vPlan} generation and \acrshort{SVA} generation is highlighted in Listing 2. The generated properties are evaluated for correctness by the critic agents and feedback is provided to improve \acrshort{SVA}. This iterative process continues until a threshold is reached without a conclusion on the \acrshort{SVA}. This is when human intervention is required to decide the correct \acrshort{SVA} and continue the overall process. The generated properties are then proven in a formal tool, and the \acrshort{CEXs} are analyzed if any and fixed by the \acrshort{LLM} agents. Once all properties are proven, another \acrshort{LLM} agent takes over to analyze the formal coverage, an important verification sign-off criteria. Based on missing coverage, feedback is provided to the formal verification lead to add missing properties.

\begin{tcolorbox}[colback=black!2!white,colframe=black,title=\small Listing 1: Formal verification lead agent example,label=fv_lead]
\footnotesize
\begin{minted}{yaml}
formal_verification_lead:
  role: >
    Formal Verification Lead
  goal: >
    Gather all the necessary information regarding the given Register Transfer Level (RTL) design and
    its specification to define a set of formal properties to verify its functionality.
  backstory: >
    An expert formal verification engineer, who spends all day and night thinking about how to verify
    the given Design Under Verification (DUV) based on its specification. It analyzes the design
    specification and defines the set of SystemVerilog Assertion (SVA) properties required to verify
    the functionality of the given design. The properties are described in the natural English
    language.
  allow_delegation: false
  verbose: true
  max_iter: 5
\end{minted}
\end{tcolorbox}

\begin{tcolorbox}[colback=black!2!white,colframe=black,title=\small Listing 2: Example of tasks defined for formal verification lead and the \acrshort{SVA} generation responsible engineer,label=fv_lead_task]
\footnotesize
\begin{minted}{yaml}
vplan_gen :
  description: >
    Analyze the specification of the Register Transfer Level (RTL) design to come up with a set of
    properties written in the natural English language that would be used to verify the functionality
    of the design using formal verification.
  expected_output: >
    SystemVerilog Assertion (SVA) properties defined in the natural English language that will be used
    to verify the functionality of the given design using formal verification.

property_gen :
  description: >
    Analyze the given property description and write the SystemVerilog Assertion (SVA) property.
  expected_output: >
   Correct SystemVerilog Assertion (SVA) for each of the properties.
\end{minted}
\end{tcolorbox}

Since \acrshort{ADHD}, hallucination and being stuck in iterative loops are the usual downsides of \acrshort{LLMs}, we circumvent these issues by dividing complete verification into smaller tasks. We also make sure that if there is a feedback loop between two agents, say, to generate and update the \acrshort{SVA}, and the loop cannot decide on the correct \acrshort{SVA}, we have set a threshold of 5 iterations, after which the human would be asked to intervene and provide the correct \acrshort{SVA} to move forward. We call this approach \say{\acrfull{HIL} \acrshort{AI}} that uses human feedback to ensure the entire flow is exercised and does not get stuck at any point. It plays a key role in making the models more truthful and reducing hallucination errors and allows human feedback to steer agents in the right direction, specify goals, and others. The \acrshort{HIL} component sits in front of the auto-reply components. It can intercept the incoming messages and decide whether to pass them to the auto-reply components or to provide human feedback based on customization in the agentic \acrshort{AI} frameworks. Fig.~\ref{hil} illustrates such a design. The algorithm for \acrshort{HIL} \acrshort{AI} is highlighted in algorithm \ref{hitl_algo}.

\begin{minipage}{0.63\textwidth}
\begin{algorithm}[H]\small
\caption{\acrfull{HIL} message processing in Saarthi}
\label{hitl_algo}
\begin{algorithmic}[1]
\REQUIRE $M$: Messages, $mode \in \{NEVER, TERMINATE\}$, $max\_replies \in N^+$
\STATE Initialize $counter \leftarrow 0$, $conversation\_active \leftarrow true$
\WHILE{conversation\_active}
    \STATE $m_t \leftarrow$ ReceiveMessage()
    \IF{IsTerminationMessage($m_t$)} 
        \STATE conversation\_active $\leftarrow false$
    \ELSIF{mode = NEVER}
        \STATE ProcessAutoReply($m_t$)
    \ELSIF{mode = TERMINATE}
        \IF{counter $\geq$ max\_replies}
            \STATE $h_c \leftarrow$ RequestHumanInput($m_t$)
            \IF{$h_c$ = TERMINATE}
                \STATE conversation\_active $\leftarrow false$
            \ELSIF{$h_c$ = SKIP}
                \STATE ProcessAutoReply($m_t$)
            \ELSIF{$h_c$ = INTERCEPT}
                \STATE ProcessHumanReply($m_t$)
                \STATE counter $\leftarrow 0$
            \ENDIF
        \ELSE
            \STATE ProcessAutoReply($m_t$)
            \STATE counter $\leftarrow$ counter + 1
        \ENDIF
    \ENDIF
\ENDWHILE
\ENSURE Counter resets on INTERCEPT; One reply per message
\end{algorithmic}
\end{algorithm}
\end{minipage}
\hfill
\begin{minipage}{0.37\textwidth}
        \includegraphics [width=\textwidth] {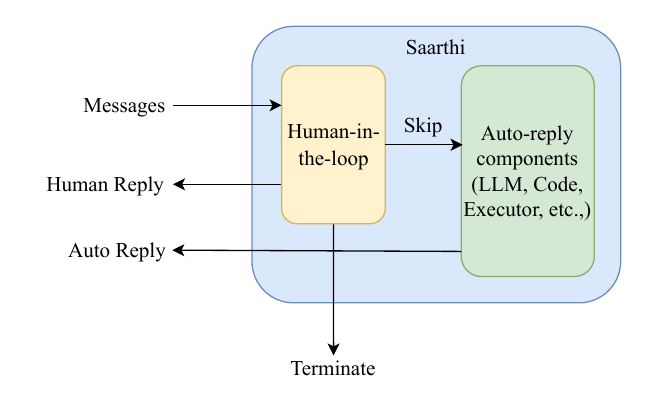}
        \captionof{figure}{Allowing human feedback in agents \cite{autogen}}
        \label{hil}
\end{minipage}

\vspace{0.4cm}
The algorithm starts by setting the conversation to active and the counter to 0. When the mode is set to NEVER, it doesn’t wait for human input and terminates automatically. However, when the mode is set to TERMINATE, it requests humans to give some input $h_c$. If $h_c$ is terminated, then the \acrshort{HIL} process terminates. If $h_c$ is skipped, then the \acrshort{HIL} process proceeds forward (e.g., to other agents) without terminating the process. If $h_c$ is intercepted, the algorithm considers and processes the human input.

\subsection{Agent Orchestration}

Once the agents are configured, they are passed as keys to an orchestration setup. This orchestration mechanism is capable of managing agents in both a sequential and hierarchical manner. For this formal verification process, the agents are arranged in a sequential order. After orchestration, the selected framework's main module initiates the verification process, invoking the agents sequentially.

During this process, the agents generate key artifacts such as \acrshort{vPlan} and properties, logging their interactions and collaborations as they proceed. The generated properties are also subjected to evaluation by critic agents, who provide feedback to improve the quality and correctness of \acrshort{SVAs}. This iterative process continues until a stable threshold is achieved. If the agents are unable to finalize the \acrshort{SVAs} within a predefined maximum number of iterations, human intervention is triggered for further assessment. Once the \acrshort{SVAs} are finalized, they undergo formal verification, with any \acrshort{CEXs} identified and resolved by the agents.

\section{Benchmarking and Results} \label{eval}

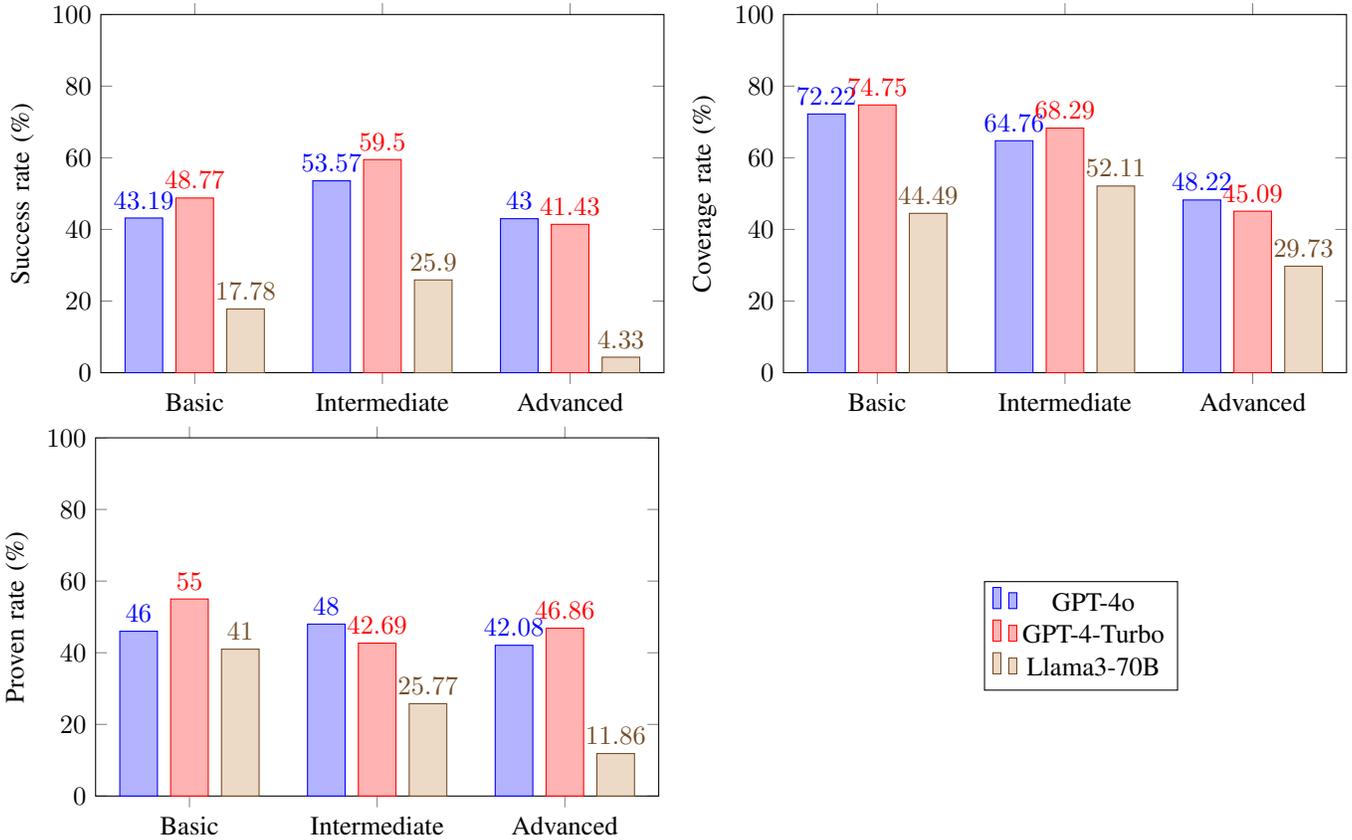
\begin{figure}[h!]
    \begin{minipage}{.5\textwidth}
        \centering
        \begin{tikzpicture}
    \begin{axis}[
            ybar=5pt,
            bar width=.5cm,
            width=\textwidth,
            height=0.7\textwidth,
            symbolic x coords={Basic,Intermediate,Advanced},
            enlarge x limits=0.25,
            xtick=data,
            nodes near coords,
            nodes near coords align={vertical},
            ymin=0,ymax=100,
            ylabel={Success rate (\%)},
        ]
        \addplot table[x=Complexity,y=GPT-4o]{\successrate};
        \addplot table[x=Complexity,y=GPT-4-Turbo]{\successrate};
        \addplot table[x=Complexity,y=Llama3-70B]{\successrate};
    \end{axis}
\end{tikzpicture}
        \label{temp1}
    \end{minipage}%
    \hfill
    \begin{minipage}{0.5\textwidth}
        \centering
        \begin{tikzpicture}
    \begin{axis}[
            ybar=5pt,
            bar width=.5cm,
            width=\textwidth,
            height=0.7\textwidth,
            symbolic x coords={Basic,Intermediate,Advanced},
            enlarge x limits=0.25,
            xtick=data,
            nodes near coords,
            nodes near coords align={vertical},
            ymin=0,ymax=100,
            ylabel={Coverage rate (\%)},
        ]
        \addplot table[x=Complexity,y=GPT-4o]{\coveragerate};
        \addplot table[x=Complexity,y=GPT-4-Turbo]{\coveragerate};
        \addplot table[x=Complexity,y=Llama3-70B]{\coveragerate};
    \end{axis}
\end{tikzpicture}
        \label{temp2}
    \end{minipage}

\begin{tikzpicture}
    \begin{axis}[
            ybar=5pt,
            bar width=.5cm,
            width=0.5\textwidth,
            height=0.35\textwidth,
            legend style={at={(1.75,0.6)},
                anchor=north},
            symbolic x coords={Basic,Intermediate,Advanced},
            enlarge x limits=0.25,
            xtick=data,
            nodes near coords,
            nodes near coords align={vertical},
            ymin=0,ymax=100,
            ylabel={Proven rate (\%)},
        ]
        \addplot table[x=Complexity,y=GPT-4o]{\provenrate};
        \addplot table[x=Complexity,y=GPT-4-Turbo]{\provenrate};
        \addplot table[x=Complexity,y=Llama3-70B]{\provenrate};
        \legend{GPT-4o, GPT-4-Turbo, Llama3-70B}
    \end{axis}
\end{tikzpicture}
\caption{\acrshort{KPIs} for Saarthi benchmarking}
\label{kpi_results}
\end{figure}

To evaluate the performance and benchmark its capabilities, we used Saarthi to verify \acrshort{RTL} designs of varied complexity. Tables \ref{basic_results}, \ref{int_results} and \ref{adv_results} underline the performance of Saarthi on basic, intermediate, and advanced design complexity levels. We chose three \acrfull{KPIs} to benchmark the results with different \acrshort{LLMs}. The first is \say{success rate} which determines the number of successful runs (i.e., end-to-end formal verification) out of the total runs. The second \acrshort{KPI} is the coverage rate (formal coverage after end-to-end formal verification). The third \acrshort{KPI} indicates the pass rate of the assertions generated from Saarthi with each \acrshort{LLM}. The results are highlighted in Fig.~\ref{kpi_results}.

\begin{table*}[h!]
    \caption{Performance of Saarthi on basic difficulty level designs}
    \renewcommand\arraystretch{1.2}
    \begin{center}
    \begin{tabular}{cc*{3}{|ccc}}
        \toprule
        \multirow{2}*{\textbf{Design}} & \multirow{2}*{\textbf{Metric}} & \multicolumn{3}{c}{\textbf{Pass@1}} & \multicolumn{3}{c}{\textbf{Pass@2}} & \multicolumn{3}{c}{\textbf{Pass@3}}
        \\
        \cmidrule(lr){3-5} \cmidrule(lr){6-8} \cmidrule(lr){9-11}
        & & GPT-4o & GPT-4-Turbo & Llama3 & GPT-4o & GPT-4-Turbo & Llama3 & GPT-4o & GPT-4-Turbo & Llama3
        \\
        \midrule
        \multirow{3}*{Accumulator} & \# Properties & 11 & 4 & 9 & 13 & 10 & 0 & 12 & 6 & 0 \\
 & \% Proven & 45.45\% & 50\% & 44.44\% & 53.85\% & 70\% & 0\% & 33.33\% & 16.67\% & 0\% \\
 & \% Coverage & 81.82\% & 80.95\% & 77.78\% & 80\% & 84\% & 0\% & 80.77\% & 83.58\% & 0\%  \\
        \midrule
        \multirow{3}*{8-bit ALU} & \# Properties & 10 & 5 & 18 & 13 & 13 & 0 & 16 & 14 & 0 \\
          & \% Proven & 20\%   & 20\%  & 27.78\% & 7.69\% & 92.31\% & 0\% & 37.50\% & 14.29\% & 0\% \\
          & \% Coverage & 93.29\%   & 94.33\%  & 90.53\% & 95.54\% & 87.59\% & 0\% & 92.68\% & 93.13\% & 0\%  \\
        \midrule
        \multirow{3}{*}{Edge detector} & \# Properties & 6       & 8       & 7       & 5    & 7       & 8    & 7       & 9       & 10      \\
                               & \% Proven    & 33.33\% & 62.50\% & 14.29\% & 40\% & 71.43\% & 75\% & 28.57\% & 44.44\% & 40\%    \\
                               & \% Coverage  & 60\%    & 60\%    & 53.85\% & 50\% & 72.73\% & 75\% & 50\%    & 73.33\% & 70.59\% \\
        \midrule
        \multirow{3}*{4-state FSM}  & \# Properties & 6 & 5 & 9 & 9 & 9 & 7 & 9 & 8 & 8 \\
 & \% Proven & 100\% & 100\% & 88.89\% & 100\% & 77.78\% & 57.14\% & 100\% & 100\% & 75\% \\
 & \% Coverage & 82.22\% & 80.95\% & 65.23\% & 52.94\% & 83.33\% & 78.74\% & 75\% & 75.59\% & 71.25\% \\
        \midrule
        \multirow{3}*{Up-down counter}  & \# Properties & 6 & 8 & 8 & 9 & 8 & 8 & 9 & 0 & 8 \\
 & \% Proven & 33\% & 75\% & 88\% & 33\% & 25\% & 25\% & 26\% & 0\% & 75\% \\
 & \% Coverage & 56\% & 69\% & 0\% & 53\% & 71\% & 79\% & 81\% & 0\% & 7\% \\
        \bottomrule
    \end{tabular}
    \label{basic_results}
    \end{center}
\end{table*}

\begin{table*}[h!]
    \caption{Performance of Saarthi on intermediate difficulty level designs}
    \renewcommand\arraystretch{1.2}
    \begin{center}
    \begin{tabular}{cc*{3}{|ccc}}
        \toprule
        \multirow{2}*{\textbf{Design}} & \multirow{2}*{\textbf{Metric}} & \multicolumn{3}{c}{\textbf{Pass@1}} & \multicolumn{3}{c}{\textbf{Pass@2}} & \multicolumn{3}{c}{\textbf{Pass@3}}
        \\
        \cmidrule(lr){3-5} \cmidrule(lr){6-8} \cmidrule(lr){9-11}
        & & GPT-4o & GPT-4-Turbo & Llama3 & GPT-4o & GPT-4-Turbo & Llama3 & GPT-4o & GPT-4-Turbo & Llama3
        \\
        \midrule
        \multirow{3}*{Sync. FIFO} & \# Properties & 20 & 12 & 12 & 22 & 13 & 7 & 6 & 9 & 13 \\
 & \% Proven & 55\% & 41.67\% & 8.33\% & 59.09\% & 23.08\% & 28.57\% & 50\% & 33.33\% & 30.77\% \\
 & \% Coverage & 91.67\% & 58.71\% & 91.66\% & 92\% & 73.08\% & 71.43\% & 50\% & 87.50\% & 84.62\% \\
        \midrule
        \multirow{3}*{FSM controller}  & \# Properties & 29 & 4 & 0 & 30 & 6 & 0 & 31 & 10 & 0 \\
 & \% Proven & 48.28\% & 75\% & 0\% & 43.33\% & 33.33\% & 0\% & 52.38\% & 20\% & 0\% \\
 & \% Coverage & 72.41\% & 42.86\% & 0\% & 68.97\% & 84.62\% & 0\% & 70\% & 73.68\% & 0\% \\
        \midrule
        \multirow{3}*{Priority encoder}  & \# Properties & 14 & 11 & 12 & 7 & 12 & 16 & 4 & 10 & 11 \\
 & \% Proven & 79\% & 55\% & 17\% & 71\% & 33\% & 38\% & 50\% & 40\% & 18\% \\
 & \% Coverage & 87\% & 82\% & 83\% & 25\% & 82\% & 87\% & 50\% & 70\% & 78\%  \\
        \midrule
        \multirow{3}*{16-bit LFSR}  & \# Properties & 5 & 5 & 6 & 6 & 5 & 6 & 5 & 5 & 0 \\
 & \% Proven & 40\% & 80\% & 67\% & 50\% & 40\% & 33\% & 40\% & 40\% & 0\% \\
 & \% Coverage & 50\% & 65\% & 44\% & 40\% & 43\% & 67\% & 40\% & 71\% & 0\% \\
        \midrule
        \multirow{3}*{CRC generator}  & \# Properties & 8 & 6 & 7 & 4 & 7 & 5 & 8 & 6 & 10 \\
 & \% Proven & 25\% & 33\% & 46\% & 50\% & 43\% & 40\% & 38\% & 50\% & 60\% \\
 & \% Coverage & 80\% & 67\% & 37\% & 50\% & 57\% & 63\% & 67\% & 67\% & 75\% \\
        \bottomrule
    \end{tabular}
    \label{int_results}
    \end{center}
\end{table*}

\begin{table*}[h!]
    \caption{Performance of Saarthi on advanced difficulty level designs}
    \renewcommand\arraystretch{1.2}
    \begin{center}
    \begin{tabular}{cc*{3}{|ccc}}
        \toprule
        \multirow{2}*{\textbf{Design}} & \multirow{2}*{\textbf{Metric}} & \multicolumn{3}{c}{\textbf{Pass@1}} & \multicolumn{3}{c}{\textbf{Pass@2}} & \multicolumn{3}{c}{\textbf{Pass@3}}
        \\
        \cmidrule(lr){3-5} \cmidrule(lr){6-8} \cmidrule(lr){9-11}
        & & GPT-4o & GPT-4-Turbo & Llama3 & GPT-4o & GPT-4-Turbo & Llama3 & GPT-4o & GPT-4-Turbo & Llama3
        \\
        \midrule
        \multirow{3}*{Booth multiplier} & \# Properties & 10 & 9 & 9 & 16 & 4 & 10 & 24 & 6 & 9 \\
 & \% Proven & 30\% & 44.44\% & 22.22\% & 25\% & 50\% & 50\% & 33\% & 33.33\% & 22.22\% \\
 & \% Coverage & 43\% & 77.78\% & 77.78\% & 73\% & 42.86\% & 90\% & 65\% & 66.67\% & 88.89\% \\
        \midrule
        \multirow{3}*{Pipelined adder}  & \# Properties & 13 & 7 & 7 & 10 & 10 & 0 & 14 & 7 & 0 \\
 & \% Proven & 31\% & 42.86\% & 28.57\% & 40\% & 40\% & 0\% & 21\% & 85.71\% & 0\% \\
 & \% Coverage & 68\% & 57.14\% & 83.33\% & 44\% & 70\% & 0\% & 31\% & 13.33\% & 0\% \\
        \midrule
        \multirow{3}*{RV32I core}  & \# Properties & 12 & 14 & 10 & 14 & 14 & 12 & 14 & 14 & 10 \\
 & \% Proven & 50\% & 44.43\% & 30.02\% & 55\% & 54.24\% & 35\% & 70\% & 75\% & 44.32\% \\
 & \% Coverage & 80\% & 82\% & 50\% & 82\% & 82\% & 51\% & 82\% & 82\% & 53\%\\
        \midrule
        \multirow{3}*{PID Controller}  & \# Properties & 5 & 2 & 0 & 1 & 4 & 0 & 5 & 2 & 0 \\
 & \% Proven & 80.30\% & 50\% & 0\% & 100\% & 25\% & 0\% & 20\% & 50\% & 0\% \\
 & \% Coverage & 46.77\% & 13.60\% & 0\% & 9.40\% & 17.69\% & 0\% & 44.53\% & 42.97\% & 0\% \\
        \midrule
        \multirow{3}*{Round robin arbiter}  & \# Properties & 10 & 7 & 7 & 12 & 6 & 0 & 26 & 4 & 0 \\
 & \% Proven & 50\% & 57\% & 29\% & 42\% & 33\% & 0\% & 22\% & 25\% & 0\% \\
 & \% Coverage & 87\% & 50\% & 67\% & 73\% & 89\% & 0\% & 86\% & 78\% & 0\% \\
        \bottomrule
    \end{tabular}
    \label{adv_results}
    \end{center}
\end{table*}

\pgfplotstableread[row sep=\\,col sep=&]{
Complexity   & GPT-4o & GPT-4-Turbo & Llama3-70B \\
Basic        & 43.19  & 48.77       & 17.78 \\
Intermediate & 53.57  & 59.50       & 25.90 \\
Advanced     & 43.00  & 41.43       & 4.33  \\
}\successrate

\pgfplotstableread[row sep=\\,col sep=&]{
Complexity   & GPT-4o & GPT-4-Turbo & Llama3-70B \\
Basic        & 72.22  & 74.75       & 44.49 \\
Intermediate & 64.76  & 68.29       & 52.11 \\
Advanced     & 48.22  & 45.09       & 29.73 \\
}\coveragerate

\pgfplotstableread[row sep=\\,col sep=&]{
Complexity   & GPT-4o & GPT-4-Turbo & Llama3-70B \\
Basic        & 46.00  & 55.00       & 41.00 \\
Intermediate & 48.00  & 42.69       & 25.77 \\
Advanced     & 42.08  & 46.86       & 11.86 \\
}\provenrate

Saarthi performs the best with the GPT-4o model and the worst with the Llama3-70B model. GPT-4o has a consistent proof rate and the most consistent coverage metrics. Even though the proven rate of the assertions is lower for GPT-4o, it yields a higher overall coverage. GPT-4-Turbo has the highest proven rate variability (i.e., less consistent results with varied design complexity). It has a higher average assertion proven rate but lower overall coverage. Llama3-70B has consistently lower success rates with the highest number of attempts needed to complete the overall end-to-end formal verification. It is worth noting that this model often generates more assertions per run compared to the other two.

\acrshort{LLMs} have context length and input token restrictions that can lead to truncated critical information in large \acrshort{RTL} designs, resulting in inaccurate assertions. The quality of \acrshort{LLM}-generated outputs depends on the prompts, yet models may still deviate from guidelines, producing incorrect assertions. Selecting the optimal temperature parameter for assertions varies across designs; incorrect settings cause overly deterministic or random outputs. Robust models like GPT-4o can err by introducing implicit clocks and resets in smaller circuits. \acrshort{LLMs} often misinterpret intricate RTL constructs, generate syntactically incorrect or semantically irrelevant assertions, and produce vacuous passes in formal properties. GPT-4o can create overly complex assertions with redundant conditions, while Llama models frequently have syntax errors. Additionally, GPT-4 models may introduce unnecessary assertions and signals not present in the original \acrshort{RTL} specification.

\section{Conclusion} \label{conc}

The paper outlines a fully autonomous \acrshort{AI}-based formal verification engineer, Saarthi. Saarthi understands design specifications, creates a verification plan, assigns tasks to several \acrshort{AI} verification engineers, and communicates with formal verification tools such as Cadence Jasper to prove properties. It also analyzes \acrshort{CEXs}, assesses formal coverage, and reacts to improve it by adding missing properties for the final sign-off. Although the results for end-to-end formal verification do not yield a \SI{100}{\percent} guarantee with every run, Saarthi performs significantly well most of the time, with an overall efficacy of around \SI{40}{\percent}. The quality of the results also depends on the \acrshort{LLM} used, with \acrshort{GPT}-4o outperforming other models. As predicted in \cite{situational_awareness} and \cite{agi_spark}, achieving \acrfull{AGI} by 2027 is strikingly plausible, and we believe that Saarthi could be pivotal in reaching that milestone within the hardware design verification domain. Saarthi is based on the agentic workflow, which makes it highly scalable and adaptable to other domains such as \acrshort{RTL} design, \acrshort{UVM} testbench generation, and more. To expand its capabilities, we need to define the agents and their responsibilities to ensure they work together to solve given tasks. In the future, we will continue our experiments to extend Saarthi's applications beyond formal verification, exploring its potential in additional usecases.

\section*{Acknowledgement}
This work has been developed in the project VE-VIDES (project label 16ME0243K) which is partly funded within the Research Programme ICT 2020 by the German Federal Ministry of Education and Research (BMBF).

\printbibliography[heading=bibintoc]

\end{document}